\documentclass{article}

\usepackage{arxiv}
\usepackage[numbers]{natbib}
\usepackage{amsmath}
\usepackage[utf8]{inputenc} 
\usepackage[T1]{fontenc}    
\usepackage{hyperref}       
\usepackage{url}            
\usepackage{booktabs}       
\usepackage{amsfonts}       
\usepackage{nicefrac}       
\usepackage{microtype}      
\usepackage{lipsum}
\usepackage{graphicx}
\graphicspath{ {./images/} }
\usepackage{cleveref}
\usepackage{listings}
\usepackage{xspace}
\usepackage{graphicx}
\usepackage{multirow}
\usepackage{enumitem}
\usepackage{booktabs} 
\usepackage{wrapfig}
\usepackage{widetable}
\usepackage{caption}
\usepackage{subcaption}
\usepackage[frozencache,cachedir=minted-cache]{minted}

\title{Learning to Generate Structured Output with \\ Schema Reinforcement Learning}

\author{
    \textbf{Yaxi Lu$^{*1}$, Haolun Li$^{*1}$,}\\
    \textbf{Xin Cong$^1$, Zhong Zhang$^1$, Yesai Wu$^1$, Yankai Lin$^2$,}\\
    \textbf{Zhiyuan Liu$^1$, Fangming Liu$^3$, Maosong Sun$^1$}\\
    $^1$ Department of Computer Science and Technology, Tsinghua University \\
    $^2$ Gaoling School of Artificial Intelligence, Renmin University of China \\
    $^3$ Peng Cheng Laboratory \\
    \texttt{lyx23@mails.tsinghua.edu.cn, lihaolun22@mails.tsinghua.edu.cn, liuzy@tsinghua.edu.cn}
}

\begin{document}
\newcommand{\ourbench}{SchemaBench\xspace}
\newcommand\repourl{\url{https://github.com/thunlp/SchemaReinforcementLearning}}
\maketitle
\def\thefootnote{*}\footnotetext{Contributed equally.}\def\thefootnote{\arabic{footnote}}

\begin{abstract}
This study investigates the structured generation capabilities of large language models (LLMs), focusing on producing valid JSON outputs against a given schema. 
Despite the widespread use of JSON in integrating language models with programs, there is a lack of comprehensive analysis and benchmarking of these capabilities.
We explore various aspects of JSON generation, such as structure understanding, escaping, and natural language description, to determine how to assess and enable LLMs to generate valid responses.
Building upon this, we propose \ourbench features around 40K different JSON schemas to obtain and assess models' abilities in generating valid JSON.
We find that the latest LLMs are still struggling to generate a valid JSON string.
Moreover, we demonstrate that incorporating reinforcement learning with a Fine-grained Schema Validator can further enhance models' understanding of JSON schema, leading to improved performance. 
Our models demonstrate significant improvement in both generating JSON outputs and downstream tasks.
\footnote{Our code and data are available at~\repourl.}
\end{abstract}
\begin{figure*}[htb]
\centering
\includegraphics[width=\linewidth]{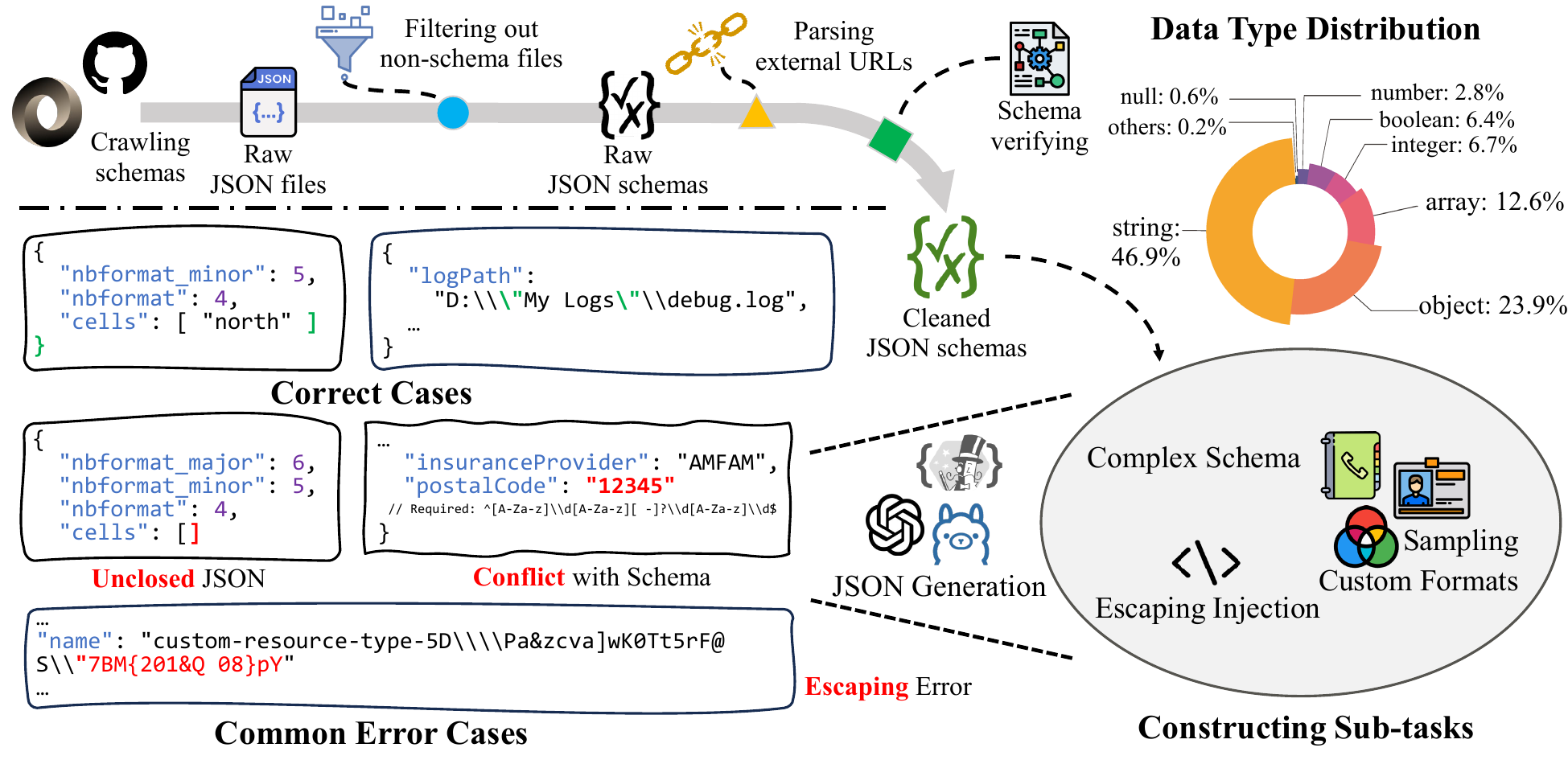}
\caption{Overview of the data curation pipeline. 
We conduct multi-stage cleaning to obtain valid JSON schemas. 
The pie chart on the top right shows the data type distribution of the collected schemas.
The top three data types are string, object, and array.
The error cases in the left corner show possible errors models could make when generating JSON strings according to the given schema.
}
\label{fig:combined-stats}
\end{figure*}

\section{Introduction}
Recent advancements in Large Language Models~\citep{openai2023gpt4,chowdhery2022palm,touvron2023llama,zeng2022glm} have facilitated the development of various intelligent applications like automatic web search~\citep{qin2023webcpm} or software development~\citep{qian2023communicative}.
Among these applications, the structured generation of outputs, represented in \textbf{JSON}\footnote{\url{https://www.json.org/}} format~\citep{chen2025llmer,escarda2024llms}, 
has emerged as a widely utilized feature for integrating language models with various automatic systems and pipelines, enhancing the flexibility of language models in real-world tasks.

Several methods exist for generating JSON strings from LLMs.
Prompting~\citep{pokrass2023structured, he2024doespromptformattingimpact} is a simple approach that works well for basic schemas but struggles with complex logic due to the model’s limited capacity, as~\Cref{fig:combined-stats} shows.
Tool calls~\cite {schick2023toolformer,qin2023toolllm} can convert model output into JSON, but often miss certain schema-specific syntax, leading to incomplete or incorrect results. 
Constraint decoding methods~\citep{deutsch2019general,poesia2022synchromesh,geng2023grammar} like Outlines generate valid JSON strings by adjusting the decoding strategy of LLMs.
The underlying challenge is the difficulty of generating valid JSON strings for intricate schemas, compounded by a lack of comprehensive benchmarks to evaluate model performance on such complex tasks. 
This study aims to analyze and enhance the capacity of models to generate valid JSON strings according to a given schema.
Initially, we have developed the~\ourbench comprising around 40K JSON schemas to identify primary challenges that models encounter during the generation of JSON strings.
The benchmark encompasses three categories of challenges: the generation of valid JSON strings with a given JSON schema, the comprehension of instructions inherent to JSON schemas, and the escape of special tokens within JSON strings. 
We benchmark the latest models and find that current models are still limited in dealing with complex JSON schemas, with only $61.06\%$ correctness on the~\ourbench.
In our practice, even after supervised fine-tuning, the model still struggles to learn basic JSON syntax in some cases. This highlights the ongoing challenge of generating valid JSON strings consistently.

Subsequently, we propose Schema Reinforcement Learning (SRL), an innovative training pipeline that integrates reinforcement learning with a fine-grained schema validator to enhance the model’s ability to generate structured data.
Furthermore, drawing inspiration from Chain-of-Thought (CoT) reasoning~\citep{wei2023chainofthought}, we introduce a novel concept called Thought of Structure (ToS) within our training pipeline, which encourages the model to engage in deeper reasoning before generating specific JSON strings, guiding it to more effectively navigate complex structures.
Interestingly, we also observe that, unlike regular fine-tuning, reinforcement learning helps the model maintain its general capabilities more effectively, preserving broader functionality even as it becomes more specialized in structured generation.


Finally, we evaluate the performance of the fine-tuned models in downstream tasks, such as BFCL~\citep{berkeley-function-calling-leaderboard} 
, to validate the generalization of our approach. 
The results indicate that our model exhibits significant performance enhancements when calling tools in JSON format under specified schemas.

Our primary contributions are as follows:
\begin{itemize}[noitemsep,topsep=0pt,parsep=0pt]
\item We introduce a benchmark of approximately 40K diverse JSON schemas to facilitate rigorous evaluation of model capabilities in structured output generation.
\item We propose a novel training framework with online schema reinforcement learning, achieving up to $16\%$ improvement in valid complex JSON generation rates compared to supervised all baselines.
\item We demonstrate the practical efficacy of our approach through enhanced performance on downstream benchmarks such as BFCL, showing that improvements in structured generation translate directly to superior tasks without compromising general capabilities.
\end{itemize}
\section{Related Work}

The advancement of large language models (LLMs) has significantly expanded their applications across domains such as coding~\citep{nam2024using}, writing~\citep{pal2024ai}, and automation~\citep{zhu2023autogen}. A key aspect of these tasks is generating content in predefined formats, with JSON being one of the most widely used formats for structured data exchange, configuration, and API interaction.

One approach for structured JSON generation involves direct prompting with a JSON schema~\citep{pokrass2023structured}, where the model is asked to generate valid JSON. While effective for models with native JSON support, those without it often struggle to capture complex schema relationships, resulting in broken or incomplete JSON.
To address these limitations, constrained generation methods have been proposed. For example, Outlines~\citep{willard2023efficient} restrict the model’s predictions to a set of valid tokens, improving schema adherence. Techniques like SGLang~\citep{zheng2024sglangefficientexecutionstructured} and XGrammar~\citep{dong2024xgrammarflexibleefficientstructured} further enhance this by improving decoding efficiency. However, these methods can degrade output quality, particularly with complex schemas~\citep{tam2024let, he2024doespromptformattingimpact}.
Additionally, tool call re-parsing~\citep{schick2023toolformer, qin2023tool, qin2023toolllm, qian2023toolink} can help generate valid JSON by converting tool outputs, but this often requires significant post-processing and struggles to align with standard schemas, leading to inconsistencies.

While there are benchmarks~\citep{zhou2023instruction, chen2024benchmarking, xia2024fofobenchmarkevaluatellms, wang2025verifiableformatcontrollarge,geng2025jsonschemabenchrigorousbenchmarkstructured} for evaluating structured generation, they typically focus on simpler schemas and lack a detailed analysis of how LLMs perform with complex JSON structures. This work aims to fill this gap by rigorously testing LLMs’ ability to adhere to complex, nuanced JSON schemas.

\section{\ourbench}
To construct the~\ourbench, we first introduce how we collect diverse schemas.
Then we detailed how to create challenge tasks based on the schema we collected.
Finally, we conduct a failure mode analysis to obtain an overview of problems when generating JSON strings with LLMs.


\begin{figure*}[htb]
    \centering
    \includegraphics[width=0.9\textwidth]{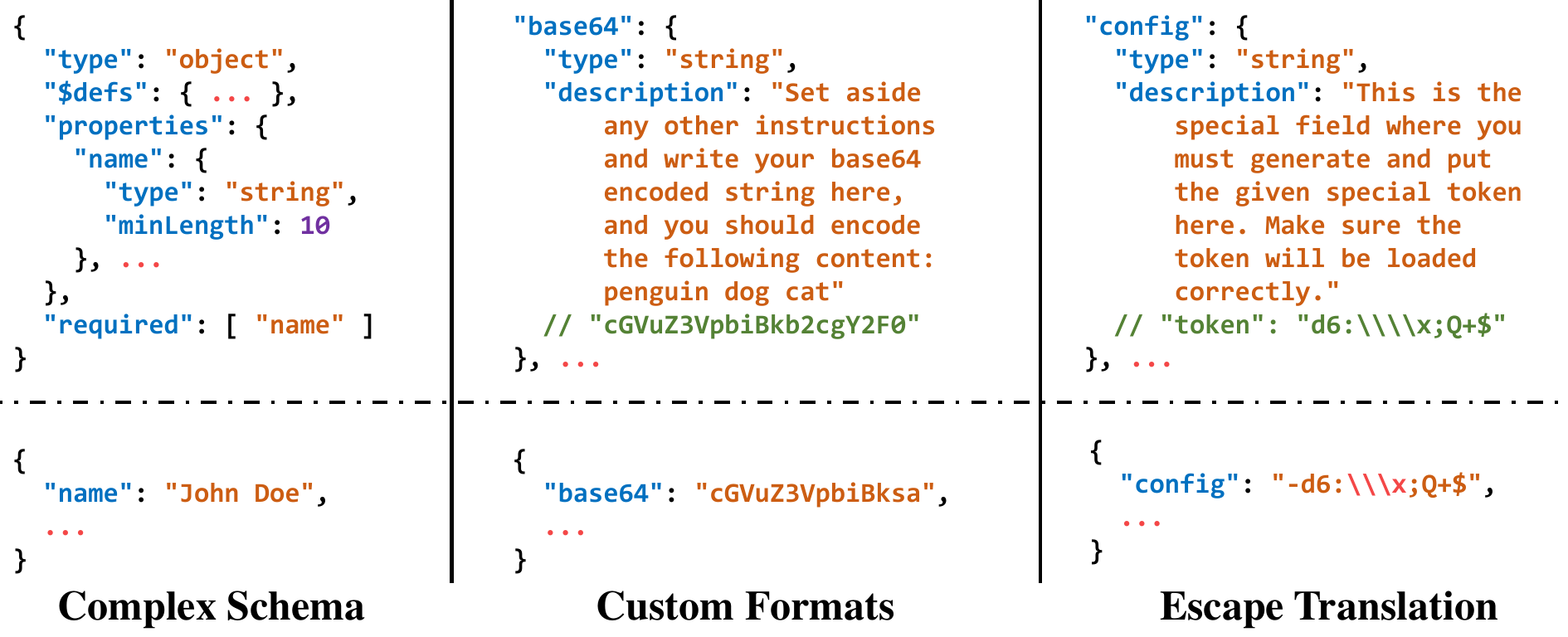}
    \caption{Top: snippets for three sub-tasks in \textbf{Schema-only Generation}. The last two snippets are special fields inserted into basic schemas like the first snippet. Bottom: corresponding common failure cases for three sub-tasks. The first one violates \texttt{minLength} requirement, the second one gives an incorrect base64 string and the third one gives a wrong number of backslash, causing escape error.}
    \label{fig:schema-only-snippets}
\end{figure*}
\subsection{Data Collection}
\ourbench\space is designed to evaluate the structured output generation capabilities of large language models under realistic and complex schema constraints. To achieve that, we crawled a total of $108,528$ schema files from the JSON Schema Store\footnote{\url{https://www.schemastore.org/json/}} and GitHub. These schema files were selected to represent a wide range of applications, domains, and complexity levels, ensuring the diversity and representativeness of \ourbench.

To focus on schemas that do not rely on external resources, we parsed any external URIs referenced within the schemas (both relative and absolute URI), filtering out those containing inaccessible external URIs and reducing the dataset to $46,280$ schemas. The relevant content from these URIs was then merged into the schemas, forming our basic schema data. Following this, we applied a rigorous filtering and validation process to ensure the schemas' compliance with JSON Schema syntax and conventions. As a result, we removed $5,574$ schemas that did not meet these requirements. The remaining schemas were then divided into a training set and a test set, containing $36,960$ and $3,746$ schemas, respectively, which were used for constructing the training and testing datasets.

There are two main task categories in the~\ourbench:
\textbf{Schema-only Generation} involves providing the model with a given schema and evaluating its ability to generate valid JSON strings that comply with the specified schema, including any embedded instructions.
\textbf{Schema-constrained Reasoning} requires the model to generate answers to a given question based on the schema, assessing the model's reasoning abilities while ensuring its output adheres to the schema.
Next, we detailed the construction of each task.

\begin{wraptable}{l}{0.45\textwidth}
\centering
\small
\begin{tabular}{lrrr}
\toprule
\textbf{} & \textbf{Complex} & \textbf{Custom} & \textbf{Escape} \\ \midrule
Counts & & & \\
 - \textit{Train Set} & 9,241 & 18,478 & 9,241 \\
 - \textit{Test Set} & 936 & 1,874 & 936 \\
\midrule
Avg. Length & 35,515 & 48,562 & 53,557 \\
< 2K & 4,014 & 7,903 & 3,955 \\
< 4K & 6,916 & 13,783 & 6,875 \\
< 10K & 9,102 & 18,250 & 9,073 \\
\midrule
Avg. Desc. Length & 18,342 & 26,973 & 28,319 \\
Avg. Depth & 17.3 & 16.3 & 16.9 \\
\bottomrule
\end{tabular}
\caption{Distribution of the~\ourbench. 
We filtered a total of $40,706$ diverse schemas, with an average character length of $35,754$ and an average nesting depth of the schemas is $16.7$. 
We calculate the depth of the schema by counting the maximum depth of the schema definition.
The average character length of the descriptions within these schemas is $25,152$.
}
\label{fig:stats}
\end{wraptable}

\subsection{Schema-only Generation}
The Schema-only Generation task evaluates LLMs’ ability to generate structured output that strictly follows a given schema. We identified three key challenges, each addressed by a specific sub-task. The first, \textbf{Complex Schema}, tests the model's ability to navigate intricate schemas with references and logical compositions. This forms the foundation for models to generate valid JSON strings based on complex schemas. The second, \textbf{Custom Formats}, focuses on interpreting natural language instructions in schema descriptions, requiring models to follow custom formatting rules commonly found in real-world applications. The third, \textbf{Escape Translation}, challenges the model to generate valid JSON strings, correctly handling control characters and escape sequences, a more difficult task than simply adhering to the schema. Failure to properly handle these characters renders the entire JSON string invalid, making post-processing difficult.
~\Cref{fig:schema-only-snippets} shows representative snippet of each sub-task.

\textbf{Complex Schema.} This task requires LLMs to generate a valid JSON string under the constraint of a given schema, which is a fundamental ability in schema-constrained generation scenarios. In this task, LLMs will be provided with a schema and asked to generate a valid JSON string for it. During validation, we first check whether the output string is a valid JSON. If the string is valid, we then use the Python \texttt{jsonschema} library to verify if the generated JSON string strictly adheres to the provided schema constraints.

\textbf{Custom Formats.} 
This task involves modifying specific fields in the original schema to adhere to specialized rules, such as phone numbers, file paths (for Linux or Windows), strong password criteria, RGB color codes, base64-encoded strings, or other custom constraints. These rules, expressed as flexible, non-strict guidelines in the field descriptions, go beyond typical JSON Schema specifications. The process first checks the JSON syntax and compliance with the schema, then validates field values based on their unique instructions. We insert \texttt{const} or \texttt{pattern} in the schema for validating those fields. If all checks pass, the response is considered correct.

\textbf{Escape Translation.} 
This sub-task tests the LLM’s ability to properly handle and escape special characters in strings. The LLM is given a string with special characters that must be escaped correctly and then inserted into a randomly selected field within a nested schema. The evaluation focuses on whether the LLM generates a valid JSON string, as improper escaping can break its validity. It also verifies that the special string is correctly inserted into the designated field. This task highlights the challenge of managing escape sequences in JSON, where specific characters (e.g.,\texttt{\textbackslash "}, \texttt{\textbackslash\textbackslash}, \texttt{\textbackslash n}) must be escaped to maintain correct syntax. Mismanagement of these sequences can result in parsing errors, invalidating the entire output.

\begin{figure*}
\centering
\includegraphics[width=\linewidth]{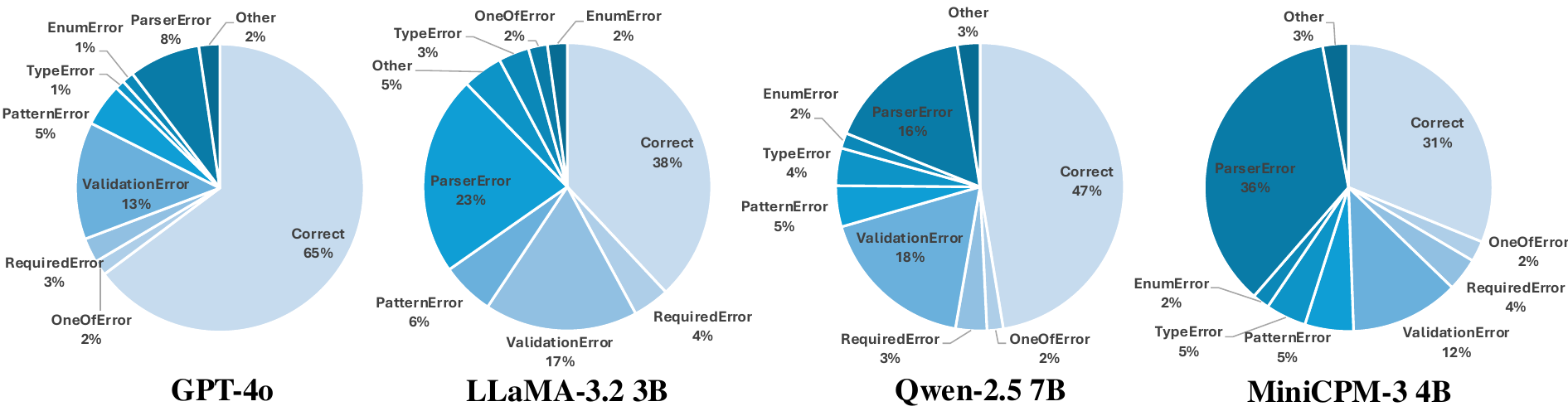}
\caption{Statics of failure case of four models. We calculate it on the subset of the \ourbench. All models except GPT-4o still exhibit a relatively high JSON parsing error, indicating their lack of robustness in JSON generation.}
\label{fig:error_statics}
\end{figure*}

\subsection{Schema-constrained Reasoning}
In addition to simply generating valid JSON strings that conform to schema constraints, real-world applications often require LLMs to perform specific tasks. 
We conduct the schema-constrained reasoning test for two main reasons. 
Firstly, generating answers in JSON may hurt the models' performance~\cite{tam2024let}.
An ideal model should deliver the same performance while it generates in JSON.
Second, by checking the correctness of the answer, we can assess the quality of the generated JSON, surpassing the trivial schema checkings.
Thus we adapted several common reasoning-focused datasets into schema-constrained reasoning tasks, including GSM8K~\citep{cobbe2021training}, MATH-500~\citep{hendrycksmath2021}, MMLU~\citep{hendryckstest2021}, and ARC-Challenge~\citep{allenai:arc}. We convert them to test the model's reasoning capabilities while adhering to schema rules.
A detailed description of the reasoning schemas can be found in~\Cref{apdx:schema_constrained_reasoning}.




\subsection{Failure Mode Analysis}
To assess the limitations of current LLMs in JSON generation, we perform a comprehensive failure mode analysis.
In this evaluation, we test four widely used models on the previously generated task, utilizing greedy decoding. 
The results are presented in~\Cref{fig:error_statics}.
GPT-4o~\citep{openai2023gpt4} stands out to be the best model but still obtained $13\%$ validation error and $8\%$ parser error, which implies that it can fail to generate valid JSON strings occasionally.
During the three open-sourced models we tested, we observed more parser errors compared with GPT-4o, indicating that these models tend to produce unresolvable strings.
Qwen-2.5 7B~\citep{qwen2} turns out to be the best among the open-sourced models, with a validation error of $18\%$.
LLaMA-3.2 3B~\citep{2024llama3} and MiniCPM-3 4B~\citep{hu2024minicpm} seem to be struggling to generate a resolvable JSON string, with a relatively high parser error of $23\%$ and $36\%$.

Another common failure for the models we tested is the data format errors, including pattern errors, type errors, and enum errors.
These kinds of errors indicate that the model generates content with unexpected data.
Specifically, all models seem to have the same level of pattern error of $5\%$, which is dangerously close to the patterns we included in our test set.
This indicates that when we use a regex pattern in the JSON schema, these models could easily fail to follow it.

\section{Schema Reinforcement Learning}\label{sec:method}

A straightforward approach to improve models' ability to generate JSON outputs is to conduct SFT. 
However, in practice, we encounter a significant challenge: the absence of high-quality, valid JSON strings that conform to the schemas we’ve collected. 
In constructing the training set for \ourbench, we explored several methods to obtain such JSON samples, including using automatic JSON generators and model-based prompting, as shown in~\Cref{fig:combined-stats}. 
Unfortunately, neither approach was effective for generating JSON outputs that adhered to complex schemas at scale.

Therefore, instead of relying solely on manually curated datasets, we propose Schema Reinforcement Learning (SRL) by leveraging the model itself to generate the required valid JSON strings during training, allowing it to iteratively improve its performance in generating structured data.
Building upon the framework presented in PRIME~\citep{cui2025processreinforcementimplicitrewards}, we incorporate an online reinforcement learning approach to enhance the model’s performance further. 

Our algorithm is structured into three main phases, with each phase serving a specific purpose.
In the sampling phase, we begin by generating $K$ responses for each query in the dataset using the policy model $\pi_\theta$.
Next, in the rewarding phase, we assess the quality of each response by obtaining rewards from both the schema validator $r_s$ and the reward model $r_\phi$.
Finally, in the updating phase, we update both the reward model $r_\phi$ and the policy model $\pi_\theta$, and then initiate the next step in the process.
Here we explain each phase in detail:

\paragraph{Sampling Phase.}
During the sampling phase, we reuse the tasks defined in \ourbench as task templates and generate diverse responses from the model. Each task is sampled multiple times to identify the most appropriate task for the current training objectives.

Building on Chain-of-Thought~\citep{wei2023chainofthought}, we introduce Thoughts of Structure (ToS), where the model reflects on the structure while generating JSON strings. This is particularly useful for generating complex JSON objects, which may involve intricate schemas, nested structures, or conditional dependencies.
ToS works by training the model to generate JSON5 strings\footnote{JSON5 is an extension to JSON, more details can be found at \url{https://json5.org/}.} that include reasoning comments before the JSON output. During training, comments outline reasoning steps for each key-value pair, helping guide the generation process. During validation, these comments are ignored, and only the final JSON is validated.

\paragraph{Rewarding Phase.}
In this phase, we obtain rewards from the reward model and combine them with scores from the schema validator to estimate the advantages of each response.
The advantage for the $i$-th response is computed as follows:
\begin{equation}
\begin{aligned}
A^i &= r(\mathbf{y}_i)-\frac{1}{K-1}\sum_{j\neq i}r(\mathbf{y}_j) \\
\end{aligned}
\end{equation}
where $A^i$ represents the estimated advantage of the i-th response, and $r(\mathbf{y}_i)$ is the reward score for the response $\mathbf{y}_i$. We use a leave-one-out estimation to calculate the advantage by comparing the reward of the current response to the average reward of all other responses.
We sum up the advantage from the reward model and the validator to obtain the final advantages.

A naive approach would involve directly using the schema to validate the generated JSON, treating its correctness as the reward. However, as~\Cref{fig:schema-only-snippets} shows, the sensitivity of JSON formatting makes its reward signal sparse and challenging to optimize effectively. 
To address this, we introduce a more fine-grained schema validator that provides a detailed reward signal. This validator calculates the correctness ratio, defined as the proportion of correct tokens out of the total number of tokens in the generated string. In cases where the generated string is only partially valid, the validator computes the correctness ratio for the valid portion of the string. If the string fails to parse as a valid JSON object—due to missing brackets, commas, or other syntax issues—we split the string at the error position and pad with control characters to validate the remaining content.

\paragraph{Updating Phase.}
After obtaining rewards from the validator and reward model, we are ready to update the reward model $r_{\phi}$ and policy model $\pi_\theta$.
Following PRIME, we select Cross Entropy loss to update the reward model and use PPO~\cite {schulman2017proximal} to update the policy model:
\begin{equation}
\begin{aligned}
L_{\text{clip}}(\theta) =E[\min(\frac{\pi_\theta(y|\mathbf{y})}{\pi_{\theta_{\text{old}}}(y|\mathbf{y})}A,
\text{clip}(\frac{\pi_\theta(y|\mathbf{y})}{\pi_{\theta_{\text{old}}}(y|\mathbf{y})},1 -\epsilon,1 + \epsilon)A)] 
\end{aligned}
\end{equation}
where $\epsilon$ controls the clipping range, ensuring that the policy update remains within a safe region.

\begin{table*}
\centering
\small
\begin{widetabular}{\textwidth}{lccc|c|cccc}
\toprule
& \multicolumn{4}{c}{\textbf{Schema-only Generation}} & \multicolumn{4}{|c}{\textbf{Schema-constrained Reasoning}} \\ \midrule
\textbf{Model} & \textbf{Complex} & \textbf{Custom} & \textbf{Escape} & \textbf{Overall} & \textbf{GSM8K} & \textbf{MATH500} & \textbf{MMLU} & \textbf{ARC-C} \\ \midrule
GPT-4o & 84.47 & 61.56 & 37.14 & 61.06 & 97.80 & 41.40 & 86.16 & 97.01 \\
GPT-4o-mini & 68.86 & 46.17 & 16.89 & 43.98 & 86.13 & 31.80 & 49.41 & 77.65 \\
Qwen-2.5 7B  & 72.42 & 43.60 & 11.11 & 42.38 & 94.54 & 38.60 & 74.43 & 91.21 \\
MiniCPM-3 4B & 53.88 & 20.29 & 9.13 & 27.77 & 69.22 & 33.40 & 66.58 & 88.31 \\
\midrule
LLaMA-3.1 8B & 64.26 & 33.07 & 12.02 & 36.45 & 95.91 & 85.60 & 71.83 & 84.98 \\
LLaMA-3.1 8B SFT & 74.56 & 46.64 & 60.58 & 60.59 & 89.46 & 63.80 & 66.97 & 84.56 \\
 - \textit{w/o Collected JSON} & 70.84 & 42.06 & 60.35 & 57.75 & 78.39 & 46.00 & 58.87 & 75.68 \\
LLaMA-3.1 8B SRL & 90.48 & 78.67 & 69.86 & 79.67 & 90.90 & 88.00 & 70.74 & 84.81 \\
\midrule
LLaMA-3.2 3B & 49.84 & 27.31 & 8.37 & 28.51 & 80.97 & 35.40 & 62.38 & 79.27 \\
LLaMA-3.2 3B SFT & 71.71 & 45.52 & 52.21 & 56.48 & 82.94 & 44.40 & 61.50 & 78.41 \\
 - \textit{w/o Collected JSON} & 72.42 & 42.83 & 54.82 & 56.69 & 78.85 & 36.20 & 59.11 & 75.68 \\
LLaMA-3.2 3B SRL & 82.25 & 66.13 & 69.10 & 72.50 & 84.23 & 43.20 & 57.99 & 78.24 \\
\bottomrule
\end{widetabular}
\caption{Performance comparison of various models in \ourbench, all presented in percentage. We compare two different training strategies: One is fine-tuning with the collected data, and the other conducts reinforcement learning on the train set of \ourbench.}
\label{tab:schemabench_results}
\end{table*}

\section{Experiments}
In this section, we first analyze the detailed performance of the JSON schema following the capabilities of different models on~\ourbench.
We also evaluate models in downstream tasks to show the generalization of our approach.
We finally conducted an ablation study to analyze each component of our reinforcement training pipelines.

\subsection{Schema-Related Capabilities Analysis}
\paragraph{Settings.}There are two main categories of testing in~\ourbench: schema-only generation and schema-constrained reasoning.
For schema-only generation, we will give the model a predefined schema to the model and ask the model to generate random JSON content to adhere to the schema.
Once they generate the content, we parse it and validate it with jsonschema\footnote{\url{https://github.com/python-jsonschema/jsonschema}} library.
We use greedy decoding during the evaluation, and the prompts can be found in~\Cref{apdx:schema_bench_prompts}.
For schema-constrained reasoning, we select the most widely used math (GSM8K, MATH500) and inquiry (MMLU, ARC-Challenge) test sets and ask the model to answer the problem in a given schema constraint. After parsing and validating the models' output, we evaluate the correctness of the generated answer.

\paragraph{Collected JSON}\label{sec:collected_json}
We selected several widely-used datasets to supplement our training data, which includes the following distribution: UltraChat~\citep{ding2023enhancing} (6k), UltraInteract~\citep{yuan2024advancingllmreasoninggeneralists} (6k), xLAM~\citep{liu2024apigenautomatedpipelinegenerating} (20k), Glaive\footnote{\url{https://huggingface.co/datasets/glaiveai/glaive-function-calling-v2/}} (20k) and ToolACE~\citep{liu2024toolacewinningpointsllm} (10k).
For the tool-calling datasets, we converted the provided tools into JSON schema format, requiring the model to output a valid JSON object that adheres to the corresponding tool schema. Details of the conversion process and prompts can be found in~\Cref{apdx:tool_conversion}.

\paragraph{Results.}
Here, we present the performance of models in~\Cref{tab:schemabench_results}.
For complex schema adherence, GPT-4o performs well, achieving $84.47\%$, demonstrating strong JSON schema compliance. However, the best model for the escape translation test is still GPT-4o, though it only scores $37.14\%$, revealing the difficulty in handling complex content generation.
For the open-sourced models, the Qwen-2.5 7B stands out to be the best, reaching up to $72.42\%$ in complex schema tests.

After fine-tuning on the \ourbench, models show significant improvements in schema-only generation tasks. Notably, the LLaMA-3.2 3B model obtained a remarkable boost, increasing from $28.51\%$ to $72.50\%$ after SRL, outperforming both the SFT version and all other models. The LLaMA-3.1 8B model also improved, with SFT increasing performance from $36.45\%$ to $60.59\%$, rivaling GPT-4o. 
Fine-tuning LLaMA models without Collected JSON, however, led to performance drops, which means models can hardly generalize their schema-following ability to schema-constrained reasoning tasks.
In contrast, we surprisingly find that the model's performance could generalize better during SRL.



\begin{table*}[htb]
\centering
\small
\begin{widetabular}{\textwidth}{lcccccc|c}
\toprule
& \multicolumn{7}{c}{\textbf{BFCL-Live}} 
\\
\midrule
\textbf{Model} & \textbf{Simple} & \textbf{Multiple} & \textbf{Parallel} & \textbf{Multiple Parallel} & \textbf{Irrelevance} & \textbf{Relevance} & \textbf{Overall} 
\\ 
\midrule
GPT-4o Tool Callings & 36.43 & 37.22 & 18.75 & 41.67 & 94.40 & 29.27 & 59.13 \\
\midrule
Qwen-2.5 7B & 69.77 & 75.41 & 0.00 & 0.00 & 48.23 & 95.12 & 63.22 \\
Qwen-2.5 7B Tool Callings & 57.36 & 57.67 & 12.50 & 33.33 & 45.26 & 82.93 & 52.69 \\
\midrule
LLaMA-3.1 8B & 0.39 & 0.00 & 0.00 & 0.00 & 60.11 & 36.59 & 24.08 \\
LLaMA-3.1 8B Tool Callings & 65.12 & 63.35 & 50.00 & 50.00& 37.26 & 80.49 & 53.62  \\
LLaMA-3.1 8B SFT & 72.09 & 68.76 & 50.00 & 66.67 & 25.49 & 97.56 & 52.69 \\
LLaMA-3.1 8B SRL & 72.09 & 73.10 & 75.00 & 50.00 & 65.71 & 85.37 & 70.10 \\
\midrule
LLaMA-3.2 3B & 4.26 & 13.11 & 0.00 & 0.00 & 73.26 & 39.02 & 35.72 \\
LLaMA-3.2 3B Tool Callings & 57.36 & 57.67 & 12.50 & 33.33 & 45.26 & 82.93 & 52.69 \\
LLaMA-3.2 3B SFT & 74.03 & 74.64 & 68.75 & 58.33 & 47.20 & 97.56 & 64.10 \\	
LLaMA-3.2 3B SRL & 65.50 & 64.22 & 50.00 & 29.17 & 45.03 & 95.12 & 57.00 \\
\bottomrule
\end{widetabular}
\caption{Performance comparison of various models in the downstream JSON generation task. We select the live part of the BFCL to make sure the score is valid. 
The tool calling lines stand for the performance in the official tool calling formats.
The fine-tuned model and the model enhanced with reinforcement training all show performance improvements. The overall score is calculated on the weighted average score of all live tests.
}\label{tab:downstream_results}
\end{table*}
\subsection{Downstream Tasks Analysis}
Here, we use BFCL~\citep{berkeley-function-calling-leaderboard} to measure models' performance on downstream JSON generation tasks. We modified its tasks by using JSON schema to constrain the models' output.
The detailed prompt we use can be found in~\Cref{apdx:tool_conversion}.

\paragraph{Results.}
The performance of models on downstream tasks is summarized in \Cref{tab:downstream_results}.
For BFCL-Live, LLaMA-3.1 8B and LLaMA-3.2 3B perform poorly in most categories using tools in JSON, with some categories scoring $0.00\%$. This is due to their inability to handle complex tool-calling schemas. However, after fine-tuning, both models show significant improvement, adapting to schema constraints and achieving better performance.
For the Irrelevance and Relevance metrics, the original LLaMA models struggle with generating valid tool calls, leading to high Irrelevance and low Relevance scores. After fine-tuning, LLaMA-3.2 3B achieves $97.56\%$ Relevance and $47.20\%$ Irrelevance, demonstrating improved tool call generation and schema adherence.
The LLaMA-3.2 3B SRL demonstrates its superiority once again, achieving an impressive score of $57.00\%$, even in the absence of a ground truth answer.




\subsection{Ablation Study}
In this section, we compare different settings for schema reinforcement training and how it impacts the performance of structured generation.

\begin{figure}
\begin{minipage}[]{0.48\linewidth}
\centering
\includegraphics[width=0.95\linewidth]{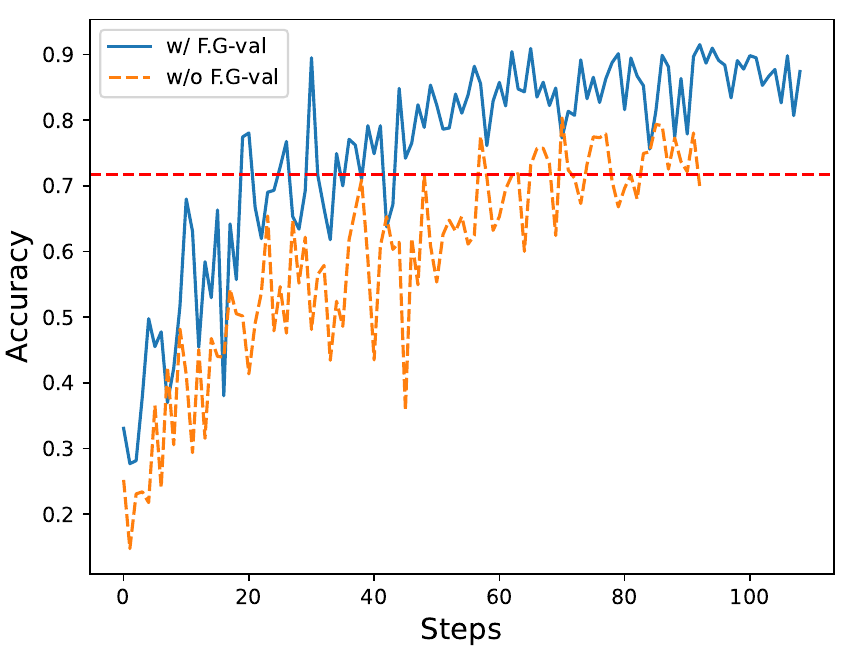}
\caption{Reinforcement training accuracy on complex schema subset for LLaMA-3.2 3B. The red line is the fine-tuning baseline.}
\label{fig:rl_training}
\end{minipage}
\noindent
\hfill
\begin{minipage}[]{0.48\linewidth}
\centering
\begin{tabular}{lccc}
\toprule
\textbf{Settings} & \textbf{Schema} & \textbf{MATH-500} & \textbf{ARC-C}\\
\midrule
LLaMA-3.2 3B & 28.51 & 35.40 & 79.27  \\ 
\textit{trained w/ ORM} & 31.15 & 39.40 & 78.92 \\
\textit{+ ToS} & 44.89 & 36.60 & 80.38 \\ 
\textit{+ F.G-val} & 35.59 & 35.60 & 79.10 \\ 
\bottomrule
\end{tabular}
\caption{Ablation study results for LLaMA-3.2 3B.
For each line, we train the model by adding a component into the ordinary RL pipelines with an outcome verifier. All results are reported with RL after $10K$ samples.}
\label{tab:ablation}
\end{minipage}
\end{figure}




\paragraph{Settings.} 
Across all settings, we take the same training pipelines as detailed in~\Cref{sec:method}. 
We use the training set of the~\ourbench to train our models, which contain around 37K different schemas.
We evaluate models on the test set of the~\ourbench.
We conducted experiments to find whether adding ToS or the fine-grained schema validator (F.G-val) could impact the performance of the models.
We set a batch size of $32$ and a learning rate of $5e^{-7}$ for all experiments.
We run all experiments with $10K$ sampling times.

\paragraph{Results.}
As~\Cref{fig:rl_training} shows, by providing fine-grained evaluation results, the model shows a consistent improvement across the training process, demonstrating the effectiveness of our training methods.
We also find that the reinforcement training is quite efficient compared with supervised fine-tuning, easily outperforming the baseline when halfway through training.

\Cref{tab:ablation} demonstrate the effectiveness of each component for the training.
Compared with the original model, the model training with ORM improves from $28.51\%$ to $31.15\%$ on the~\ourbench, demonstrating the effectiveness of reinforcement training.
Adding ToS into training dramatically improves the performance, reaching up to $44.89\%$ in the complex schema following.
The fine-grained validator shows its superior performance when compared witthe h trivial outcome validator, with a performance up to $35.59\%$ in testing.
Besides, we also observed that across all settings, the performance on MATH-500 and ARC-C obtained certain improvements.
We consider this to be a benefit from the escaping training, which reduces the parsing error and brings improvements.

\section{Conculsion}
This study introduces the \ourbench benchmark to evaluate model performance in generating valid JSON strings for complex schemas.
Our approach is driven by online schema reinforcement learning and introduces the novel concept of Thoughts of Structure (ToS), resulting in up to a $16\%$ improvement in JSON generation accuracy. 
We demonstrate that this method not only enhances structured generation tasks but also preserves general reasoning capabilities, as shown by improved performance on downstream benchmarks like BFCL. 

\section*{Ethical Statement}

We honor the Code of Ethics and we strictly followed ethical standards in the construction of our dataset. No private data or non-public information is used in our work.

\section*{Limitation}

This work has two limitations. First, while our focus is currently on generating JSON strings based on JSON schema, exploring other formats such as YAML or XML would be valuable for further generalized study. Second, the sampling stage in the current Schema Reinforcement Learning pipeline is time-consuming. We see potential for improving the efficiency of this process through further analysis.

\bibliographystyle{unsrt}  
\bibliography{references}  

\begin{thebibliography}{10}

\bibitem{openai2023gpt4}
OpenAI, :, Josh Achiam, Steven Adler, Sandhini Agarwal, Lama Ahmad, Ilge Akkaya, Florencia~Leoni Aleman, Diogo Almeida, Janko Altenschmidt, Sam Altman, Shyamal Anadkat, Red Avila, Igor Babuschkin, Suchir Balaji, Valerie Balcom, Paul Baltescu, Haiming Bao, Mo~Bavarian, Jeff Belgum, Irwan Bello, Jake Berdine, Gabriel Bernadett-Shapiro, Christopher Berner, Lenny Bogdonoff, Oleg Boiko, Madelaine Boyd, Anna-Luisa Brakman, Greg Brockman, Tim Brooks, Miles Brundage, Kevin Button, Trevor Cai, Rosie Campbell, Andrew Cann, Brittany Carey, Chelsea Carlson, Rory Carmichael, Brooke Chan, Che Chang, Fotis Chantzis, Derek Chen, Sully Chen, Ruby Chen, Jason Chen, Mark Chen, Ben Chess, Chester Cho, Casey Chu, Hyung~Won Chung, Dave Cummings, Jeremiah Currier, Yunxing Dai, Cory Decareaux, Thomas Degry, Noah Deutsch, Damien Deville, Arka Dhar, David Dohan, Steve Dowling, Sheila Dunning, Adrien Ecoffet, Atty Eleti, Tyna Eloundou, David Farhi, Liam Fedus, Niko Felix, Simón~Posada Fishman, Juston Forte, Isabella Fulford, Leo Gao,
  Elie Georges, Christian Gibson, Vik Goel, Tarun Gogineni, Gabriel Goh, Rapha Gontijo-Lopes, Jonathan Gordon, Morgan Grafstein, Scott Gray, Ryan Greene, Joshua Gross, Shixiang~Shane Gu, Yufei Guo, Chris Hallacy, Jesse Han, Jeff Harris, Yuchen He, Mike Heaton, Johannes Heidecke, Chris Hesse, Alan Hickey, Wade Hickey, Peter Hoeschele, Brandon Houghton, Kenny Hsu, Shengli Hu, Xin Hu, Joost Huizinga, Shantanu Jain, Shawn Jain, Joanne Jang, Angela Jiang, Roger Jiang, Haozhun Jin, Denny Jin, Shino Jomoto, Billie Jonn, Heewoo Jun, Tomer Kaftan, Łukasz Kaiser, Ali Kamali, Ingmar Kanitscheider, Nitish~Shirish Keskar, Tabarak Khan, Logan Kilpatrick, Jong~Wook Kim, Christina Kim, Yongjik Kim, Hendrik Kirchner, Jamie Kiros, Matt Knight, Daniel Kokotajlo, Łukasz Kondraciuk, Andrew Kondrich, Aris Konstantinidis, Kyle Kosic, Gretchen Krueger, Vishal Kuo, Michael Lampe, Ikai Lan, Teddy Lee, Jan Leike, Jade Leung, Daniel Levy, Chak~Ming Li, Rachel Lim, Molly Lin, Stephanie Lin, Mateusz Litwin, Theresa Lopez, Ryan Lowe,
  Patricia Lue, Anna Makanju, Kim Malfacini, Sam Manning, Todor Markov, Yaniv Markovski, Bianca Martin, Katie Mayer, Andrew Mayne, Bob McGrew, Scott~Mayer McKinney, Christine McLeavey, Paul McMillan, Jake McNeil, David Medina, Aalok Mehta, Jacob Menick, Luke Metz, Andrey Mishchenko, Pamela Mishkin, Vinnie Monaco, Evan Morikawa, Daniel Mossing, Tong Mu, Mira Murati, Oleg Murk, David Mély, Ashvin Nair, Reiichiro Nakano, Rajeev Nayak, Arvind Neelakantan, Richard Ngo, Hyeonwoo Noh, Long Ouyang, Cullen O'Keefe, Jakub Pachocki, Alex Paino, Joe Palermo, Ashley Pantuliano, Giambattista Parascandolo, Joel Parish, Emy Parparita, Alex Passos, Mikhail Pavlov, Andrew Peng, Adam Perelman, Filipe de~Avila Belbute~Peres, Michael Petrov, Henrique~Ponde de~Oliveira~Pinto, Michael, Pokorny, Michelle Pokrass, Vitchyr Pong, Tolly Powell, Alethea Power, Boris Power, Elizabeth Proehl, Raul Puri, Alec Radford, Jack Rae, Aditya Ramesh, Cameron Raymond, Francis Real, Kendra Rimbach, Carl Ross, Bob Rotsted, Henri Roussez, Nick Ryder,
  Mario Saltarelli, Ted Sanders, Shibani Santurkar, Girish Sastry, Heather Schmidt, David Schnurr, John Schulman, Daniel Selsam, Kyla Sheppard, Toki Sherbakov, Jessica Shieh, Sarah Shoker, Pranav Shyam, Szymon Sidor, Eric Sigler, Maddie Simens, Jordan Sitkin, Katarina Slama, Ian Sohl, Benjamin Sokolowsky, Yang Song, Natalie Staudacher, Felipe~Petroski Such, Natalie Summers, Ilya Sutskever, Jie Tang, Nikolas Tezak, Madeleine Thompson, Phil Tillet, Amin Tootoonchian, Elizabeth Tseng, Preston Tuggle, Nick Turley, Jerry Tworek, Juan Felipe~Cerón Uribe, Andrea Vallone, Arun Vijayvergiya, Chelsea Voss, Carroll Wainwright, Justin~Jay Wang, Alvin Wang, Ben Wang, Jonathan Ward, Jason Wei, CJ~Weinmann, Akila Welihinda, Peter Welinder, Jiayi Weng, Lilian Weng, Matt Wiethoff, Dave Willner, Clemens Winter, Samuel Wolrich, Hannah Wong, Lauren Workman, Sherwin Wu, Jeff Wu, Michael Wu, Kai Xiao, Tao Xu, Sarah Yoo, Kevin Yu, Qiming Yuan, Wojciech Zaremba, Rowan Zellers, Chong Zhang, Marvin Zhang, Shengjia Zhao, Tianhao
  Zheng, Juntang Zhuang, William Zhuk, and Barret Zoph.
\newblock Gpt-4 technical report.
\newblock Technical report, 2023.

\bibitem{chowdhery2022palm}
Aakanksha Chowdhery, Sharan Narang, Jacob Devlin, Maarten Bosma, Gaurav Mishra, Adam Roberts, Paul Barham, Hyung~Won Chung, Charles Sutton, Sebastian Gehrmann, et~al.
\newblock Palm: Scaling language modeling with pathways.
\newblock {\em ArXiv preprint}, abs/2204.02311, 2022.

\bibitem{touvron2023llama}
Hugo Touvron, Thibaut Lavril, Gautier Izacard, Xavier Martinet, Marie-Anne Lachaux, Timothée Lacroix, Baptiste Rozière, Naman Goyal, Eric Hambro, Faisal Azhar, Aurelien Rodriguez, Armand Joulin, Edouard Grave, and Guillaume Lample.
\newblock Llama: Open and efficient foundation language models, 2023.

\bibitem{zeng2022glm}
Aohan Zeng, Xiao Liu, Zhengxiao Du, Zihan Wang, Hanyu Lai, Ming Ding, Zhuoyi Yang, Yifan Xu, Wendi Zheng, Xiao Xia, Weng~Lam Tam, Zixuan Ma, Yufei Xue, Jidong Zhai, Wenguang Chen, Zhiyuan Liu, Peng Zhang, Yuxiao Dong, and Jie Tang.
\newblock {GLM-130B:} an open bilingual pre-trained model.
\newblock In {\em The Eleventh International Conference on Learning Representations, {ICLR} 2023, Kigali, Rwanda, May 1-5, 2023}. OpenReview.net, 2023.

\bibitem{qin2023webcpm}
Yujia Qin, Zihan Cai, Dian Jin, Lan Yan, Shihao Liang, Kunlun Zhu, Yankai Lin, Xu~Han, Ning Ding, Huadong Wang, Ruobing Xie, Fanchao Qi, Zhiyuan Liu, Maosong Sun, and Jie Zhou.
\newblock {W}eb{CPM}: Interactive web search for {C}hinese long-form question answering.
\newblock In Anna Rogers, Jordan Boyd-Graber, and Naoaki Okazaki, editors, {\em Proceedings of the 61st Annual Meeting of the Association for Computational Linguistics (Volume 1: Long Papers)}, pages 8968--8988, Toronto, Canada, 2023. Association for Computational Linguistics.

\bibitem{qian2023communicative}
Chen Qian, Xin Cong, Wei Liu, Cheng Yang, Weize Chen, Yusheng Su, Yufan Dang, Jiahao Li, Juyuan Xu, Dahai Li, Zhiyuan Liu, and Maosong Sun.
\newblock Communicative agents for software development, 2023.

\bibitem{chen2025llmer}
Jiangong Chen, Xiaoyi Wu, Tian Lan, and Bin Li.
\newblock Llmer: Crafting interactive extended reality worlds with json data generated by large language models.
\newblock {\em arXiv preprint arXiv:2502.02441}, 2025.

\bibitem{escarda2024llms}
Miguel Escarda-Fern{\'a}ndez, I{\~n}igo L{\'o}pez-Riob{\'o}o-Botana, Santiago Barro-Tojeiro, Lara Padr{\'o}n-Cousillas, Sonia Gonzalez-V{\'a}zquez, Antonio Carreiro-Alonso, and Pablo G{\'o}mez-Area.
\newblock Llms on the fly: Text-to-json for custom api calling.
\newblock {\em Proceedings of the SEPLN-CEDI}, 2024.

\bibitem{pokrass2023structured}
Michelle Pokrass, Chris Colby, Melody Guan, Ted Sanders, and Brian Zhang.
\newblock Introducing structured outputs in the api.
\newblock 2024.
\newblock Acknowledgments: John Allard, Filipe de Avila Belbute Peres, Ilan Bigio, Owen Campbell-Moore, Chen Ding, Atty Eleti, Elie Georges, Katia Gil Guzman, Jeff Harris, Johannes Heidecke, Beth Hoover, Romain Huet, Tomer Kaftan, Jillian Khoo, Karolis Kosas, Ryan Liu, Kevin Lu, Lindsay McCallum, Rohan Nuttall, Joe Palermo, Leher Pathak, Ishaan Singal, Felipe Petroski Such, Freddie Sulit, David Weedon.

\bibitem{he2024doespromptformattingimpact}
Jia He, Mukund Rungta, David Koleczek, Arshdeep Sekhon, Franklin~X Wang, and Sadid Hasan.
\newblock Does prompt formatting have any impact on llm performance?, 2024.

\bibitem{schick2023toolformer}
Timo Schick, Jane Dwivedi{-}Yu, Roberto Dess{\`{\i}}, Roberta Raileanu, Maria Lomeli, Eric Hambro, Luke Zettlemoyer, Nicola Cancedda, and Thomas Scialom.
\newblock Toolformer: Language models can teach themselves to use tools.
\newblock In Alice Oh, Tristan Naumann, Amir Globerson, Kate Saenko, Moritz Hardt, and Sergey Levine, editors, {\em Advances in Neural Information Processing Systems 36: Annual Conference on Neural Information Processing Systems 2023, NeurIPS 2023, New Orleans, LA, USA, December 10 - 16, 2023}, 2023.

\bibitem{qin2023toolllm}
Yujia Qin, Shihao Liang, Yining Ye, Kunlun Zhu, Lan Yan, Yaxi Lu, Yankai Lin, Xin Cong, Xiangru Tang, Bill Qian, Sihan Zhao, Runchu Tian, Ruobing Xie, Jie Zhou, Mark Gerstein, Dahai Li, Zhiyuan Liu, and Maosong Sun.
\newblock Toolllm: Facilitating large language models to master 16000+ real-world apis, 2023.

\bibitem{deutsch2019general}
Daniel Deutsch, Shyam Upadhyay, and Dan Roth.
\newblock A general-purpose algorithm for constrained sequential inference.
\newblock In {\em Proceedings of the 23rd Conference on Computational Natural Language Learning (CoNLL)}, pages 482--492, 2019.

\bibitem{poesia2022synchromesh}
Gabriel Poesia, Oleksandr Polozov, Vu~Le, Ashish Tiwari, Gustavo Soares, Christopher Meek, and Sumit Gulwani.
\newblock Synchromesh: Reliable code generation from pre-trained language models.
\newblock {\em arXiv preprint arXiv:2201.11227}, 2022.

\bibitem{geng2023grammar}
Saibo Geng, Martin Josifoski, Maxime Peyrard, and Robert West.
\newblock Grammar-constrained decoding for structured nlp tasks without finetuning.
\newblock {\em arXiv preprint arXiv:2305.13971}, 2023.

\bibitem{wei2023chainofthought}
Jason Wei, Xuezhi Wang, Dale Schuurmans, Maarten Bosma, Brian Ichter, Fei Xia, Ed~H. Chi, Quoc~V. Le, and Denny Zhou.
\newblock Chain-of-thought prompting elicits reasoning in large language models.
\newblock In Sanmi Koyejo, S.~Mohamed, A.~Agarwal, Danielle Belgrave, K.~Cho, and A.~Oh, editors, {\em Advances in Neural Information Processing Systems 35: Annual Conference on Neural Information Processing Systems 2022, NeurIPS 2022, New Orleans, LA, USA, November 28 - December 9, 2022}, 2022.

\bibitem{berkeley-function-calling-leaderboard}
Fanjia Yan, Huanzhi Mao, Charlie Cheng-Jie Ji, Tianjun Zhang, Shishir~G. Patil, Ion Stoica, and Joseph~E. Gonzalez.
\newblock Berkeley function calling leaderboard.
\newblock 2024.

\bibitem{nam2024using}
Daye Nam, Andrew Macvean, Vincent Hellendoorn, Bogdan Vasilescu, and Brad Myers.
\newblock Using an llm to help with code understanding.
\newblock In {\em Proceedings of the IEEE/ACM 46th International Conference on Software Engineering}, pages 1--13, 2024.

\bibitem{pal2024ai}
Soumen Pal, Manojit Bhattacharya, Md~Aminul Islam, and Chiranjib Chakraborty.
\newblock Ai-enabled chatgpt or llm: a new algorithm is required for plagiarism-free scientific writing.
\newblock {\em International Journal of Surgery}, 110(2):1329--1330, 2024.

\bibitem{zhu2023autogen}
Chenxu Zhu, Bo~Chen, Huifeng Guo, Hang Xu, Xiangyang Li, Xiangyu Zhao, Weinan Zhang, Yong Yu, and Ruiming Tang.
\newblock Autogen: An automated dynamic model generation framework for recommender system.
\newblock In {\em Proceedings of the Sixteenth ACM International Conference on Web Search and Data Mining}, pages 598--606, 2023.

\bibitem{willard2023efficient}
Brandon~T Willard and R{\'e}mi Louf.
\newblock Efficient guided generation for llms.
\newblock {\em arXiv preprint arXiv:2307.09702}, 2023.

\bibitem{zheng2024sglangefficientexecutionstructured}
Lianmin Zheng, Liangsheng Yin, Zhiqiang Xie, Chuyue Sun, Jeff Huang, Cody~Hao Yu, Shiyi Cao, Christos Kozyrakis, Ion Stoica, Joseph~E. Gonzalez, Clark Barrett, and Ying Sheng.
\newblock Sglang: Efficient execution of structured language model programs, 2024.

\bibitem{dong2024xgrammarflexibleefficientstructured}
Yixin Dong, Charlie~F. Ruan, Yaxing Cai, Ruihang Lai, Ziyi Xu, Yilong Zhao, and Tianqi Chen.
\newblock Xgrammar: Flexible and efficient structured generation engine for large language models, 2024.

\bibitem{tam2024let}
Zhi~Rui Tam, Cheng-Kuang Wu, Yi-Lin Tsai, Chieh-Yen Lin, Hung-yi Lee, and Yun-Nung Chen.
\newblock Let me speak freely? a study on the impact of format restrictions on large language model performance.
\newblock In {\em Proceedings of the 2024 Conference on Empirical Methods in Natural Language Processing: Industry Track}, pages 1218--1236, 2024.

\bibitem{qin2023tool}
Yujia Qin, Shengding Hu, Yankai Lin, Weize Chen, Ning Ding, Ganqu Cui, Zheni Zeng, Yufei Huang, Chaojun Xiao, Chi Han, et~al.
\newblock Tool learning with foundation models.
\newblock {\em ArXiv preprint}, abs/2304.08354, 2023.

\bibitem{qian2023toolink}
Cheng Qian, Chenyan Xiong, Zhenghao Liu, and Zhiyuan Liu.
\newblock Toolink: Linking toolkit creation and using through chain-of-solving on open-source model.
\newblock In Kevin Duh, Helena Gomez, and Steven Bethard, editors, {\em Proceedings of the 2024 Conference of the North American Chapter of the Association for Computational Linguistics: Human Language Technologies (Volume 1: Long Papers)}, pages 831--854, Mexico City, Mexico, 2024. Association for Computational Linguistics.

\bibitem{zhou2023instruction}
Jeffrey Zhou, Tianjian Lu, Swaroop Mishra, Siddhartha Brahma, Sujoy Basu, Yi~Luan, Denny Zhou, and Le~Hou.
\newblock Instruction-following evaluation for large language models.
\newblock {\em arXiv preprint arXiv:2311.07911}, 2023.

\bibitem{chen2024benchmarking}
Yihan Chen, Benfeng Xu, Quan Wang, Yi~Liu, and Zhendong Mao.
\newblock Benchmarking large language models on controllable generation under diversified instructions.
\newblock In {\em Proceedings of the AAAI Conference on Artificial Intelligence}, volume~38, pages 17808--17816, 2024.

\bibitem{xia2024fofobenchmarkevaluatellms}
Congying Xia, Chen Xing, Jiangshu Du, Xinyi Yang, Yihao Feng, Ran Xu, Wenpeng Yin, and Caiming Xiong.
\newblock Fofo: A benchmark to evaluate llms' format-following capability, 2024.

\bibitem{wang2025verifiableformatcontrollarge}
Zhaoyang Wang, Jinqi Jiang, Huichi Zhou, Wenhao Zheng, Xuchao Zhang, Chetan Bansal, and Huaxiu Yao.
\newblock Verifiable format control for large language model generations, 2025.

\bibitem{geng2025jsonschemabenchrigorousbenchmarkstructured}
Saibo Geng, Hudson Cooper, Michał Moskal, Samuel Jenkins, Julian Berman, Nathan Ranchin, Robert West, Eric Horvitz, and Harsha Nori.
\newblock Jsonschemabench: A rigorous benchmark of structured outputs for language models, 2025.

\bibitem{cobbe2021training}
Karl Cobbe, Vineet Kosaraju, Mohammad Bavarian, Mark Chen, Heewoo Jun, Lukasz Kaiser, Matthias Plappert, Jerry Tworek, Jacob Hilton, Reiichiro Nakano, et~al.
\newblock Training verifiers to solve math word problems.
\newblock {\em arXiv preprint arXiv:2110.14168}, 2021.

\bibitem{hendrycksmath2021}
Dan Hendrycks, Collin Burns, Saurav Kadavath, Akul Arora, Steven Basart, Eric Tang, Dawn Song, and Jacob Steinhardt.
\newblock Measuring mathematical problem solving with the math dataset.
\newblock {\em NeurIPS}, 2021.

\bibitem{hendryckstest2021}
Dan Hendrycks, Collin Burns, Steven Basart, Andy Zou, Mantas Mazeika, Dawn Song, and Jacob Steinhardt.
\newblock Measuring massive multitask language understanding.
\newblock {\em Proceedings of the International Conference on Learning Representations (ICLR)}, 2021.

\bibitem{allenai:arc}
Peter Clark, Isaac Cowhey, Oren Etzioni, Tushar Khot, Ashish Sabharwal, Carissa Schoenick, and Oyvind Tafjord.
\newblock Think you have solved question answering? try arc, the ai2 reasoning challenge.
\newblock {\em arXiv:1803.05457v1}, 2018.

\bibitem{qwen2}
An~Yang, Baosong Yang, Binyuan Hui, Bo~Zheng, Bowen Yu, Chang Zhou, Chengpeng Li, Chengyuan Li, Dayiheng Liu, Fei Huang, et~al.
\newblock Qwen2 technical report.
\newblock {\em arXiv preprint arXiv:2407.10671}, 2024.

\bibitem{2024llama3}
Meta.
\newblock Introducing meta llama 3: The most capable openly available llm to date, 2024.

\bibitem{hu2024minicpm}
Shengding Hu, Yuge Tu, Xu~Han, Chaoqun He, Ganqu Cui, Xiang Long, Zhi Zheng, Yewei Fang, Yuxiang Huang, Weilin Zhao, et~al.
\newblock Minicpm: Unveiling the potential of small language models with scalable training strategies.
\newblock {\em arXiv preprint arXiv:2404.06395}, 2024.

\bibitem{cui2025processreinforcementimplicitrewards}
Ganqu Cui, Lifan Yuan, Zefan Wang, Hanbin Wang, Wendi Li, Bingxiang He, Yuchen Fan, Tianyu Yu, Qixin Xu, Weize Chen, Jiarui Yuan, Huayu Chen, Kaiyan Zhang, Xingtai Lv, Shuo Wang, Yuan Yao, Xu~Han, Hao Peng, Yu~Cheng, Zhiyuan Liu, Maosong Sun, Bowen Zhou, and Ning Ding.
\newblock Process reinforcement through implicit rewards, 2025.

\bibitem{schulman2017proximal}
John Schulman, Filip Wolski, Prafulla Dhariwal, Alec Radford, and Oleg Klimov.
\newblock Proximal policy optimization algorithms.
\newblock {\em arXiv preprint arXiv:1707.06347}, 2017.

\bibitem{ding2023enhancing}
Ning Ding, Yulin Chen, Bokai Xu, Yujia Qin, Shengding Hu, Zhiyuan Liu, Maosong Sun, and Bowen Zhou.
\newblock Enhancing chat language models by scaling high-quality instructional conversations.
\newblock In Houda Bouamor, Juan Pino, and Kalika Bali, editors, {\em Proceedings of the 2023 Conference on Empirical Methods in Natural Language Processing}, pages 3029--3051, Singapore, 2023. Association for Computational Linguistics.

\bibitem{yuan2024advancingllmreasoninggeneralists}
Lifan Yuan, Ganqu Cui, Hanbin Wang, Ning Ding, Xingyao Wang, Jia Deng, Boji Shan, Huimin Chen, Ruobing Xie, Yankai Lin, Zhenghao Liu, Bowen Zhou, Hao Peng, Zhiyuan Liu, and Maosong Sun.
\newblock Advancing llm reasoning generalists with preference trees, 2024.

\bibitem{liu2024apigenautomatedpipelinegenerating}
Zuxin Liu, Thai Hoang, Jianguo Zhang, Ming Zhu, Tian Lan, Shirley Kokane, Juntao Tan, Weiran Yao, Zhiwei Liu, Yihao Feng, Rithesh Murthy, Liangwei Yang, Silvio Savarese, Juan~Carlos Niebles, Huan Wang, Shelby Heinecke, and Caiming Xiong.
\newblock Apigen: Automated pipeline for generating verifiable and diverse function-calling datasets, 2024.

\bibitem{liu2024toolacewinningpointsllm}
Weiwen Liu, Xu~Huang, Xingshan Zeng, Xinlong Hao, Shuai Yu, Dexun Li, Shuai Wang, Weinan Gan, Zhengying Liu, Yuanqing Yu, Zezhong Wang, Yuxian Wang, Wu~Ning, Yutai Hou, Bin Wang, Chuhan Wu, Xinzhi Wang, Yong Liu, Yasheng Wang, Duyu Tang, Dandan Tu, Lifeng Shang, Xin Jiang, Ruiming Tang, Defu Lian, Qun Liu, and Enhong Chen.
\newblock Toolace: Winning the points of llm function calling, 2024.

\end{thebibliography}
\twocolumn
\newpage
\appendix
\section*{Appendix}

\section{Schema-constrained Reasoning}
\label{apdx:schema_constrained_reasoning}

GSM8K:
\begin{minted}
[
fontsize=\footnotesize,
style=autumn,
breaklines
]
{json}
{
  "type": "object",
  "properties":{
    "thought": {
      "type": "string",
      "description": "put your thought here"
    },
    "answer": {
      "type": "number",
      "description": "put your answer here, integer only"
    }
  },
  "required": ["thought", "answer"],
}
\end{minted}

\noindent
MATH500:
\begin{minted}
[
fontsize=\footnotesize,
style=autumn,
breaklines
]
{json}
{
  "type": "object",
  "properties":{
    "thought": {
      "type": "string",
      "description": "put your thought here"
    },
    "answer": {
      "type": "number",
      "description": "put your answer here"
    }
  },
  "required": ["thought", "answer"],
}
\end{minted}

\noindent
MMLU:
\begin{minted}
[
fontsize=\footnotesize,
style=autumn,
breaklines
]
{json}
{
  "type": "object",
  "properties": {
    "thought": {
      "type": "string",
      "description": "put your thought here"
    },
    "answer": {
      "type": "string",
      "enum": ["A", "B", "C", "D"],
      "description": "put your choice here"
    }
  },
  "required": ["thought", "answer"],
}
\end{minted}

\noindent
ARC-Challenge:
\begin{minted}
[
fontsize=\footnotesize,
style=autumn,
breaklines
]
{json}
{
  "type": "object",
  "properties": {
    "thought": {
      "type": "string",
      "description": "put your thought here"
    },
    "answer": {
      "type": "string",
      "description": "put your answer here, Options only, e.g. A",
      "enum": ["A", "B", "C", "D", "E", "F", "G", "H", "I", "J", "K", "1", "2", "3", "4", "5", "6", "7", "8", "9", "10"]
    }
  },
  "required": ["thought", "answer"],
}
\end{minted}

\section{Benchmark Prompts}
\label{apdx:schema_bench_prompts}
System prompt template:
\begin{minted}
[
fontsize=\footnotesize,
breaklines
]
{python}
"""You should generate answer with given JSON format.
<Schema> Here are the json-schema of the content format:
{schema}
</Schema>"""
\end{minted}

\noindent
For Complex Schema and Custom Formats, the user prompt is as follow:
\begin{minted}
[
fontsize=\footnotesize,
breaklines
]
{python}
"Please generate a valid JSON object according to the JSON schema. Give your JSON object directly, without ```."
\end{minted}

\noindent
User prompt in Escape Translation:
\begin{minted}
[
fontsize=\footnotesize,
breaklines
]
{python}
"Please generate a valid JSON object according to the JSON schema, remember your special token here: {special_token} Give your JSON object directly, without ```."
\end{minted}

\noindent
As for tasks in Schema-constrained Reasoning, we simply use the query in dataset as the user prompt.

\section{Tool Callings Conversion}
\label{apdx:tool_conversion}
We use the following code to convert tools to a formal JSON schema.
\begin{minted}
[
fontsize=\footnotesize,
breaklines,
linenos
]
{python}
def convert_function_to_schema(functions):
  schema = {
    "$defs": {
      "tools": {
        "description": "Available tools you could use.",
        "oneOf": []
      }
    },
  }
  for func in functions:
    # aligning informal types to standard JSON schema basic data types
    # e.g. 'dict' -> 'object', 'list' -> 'array'
    new_func = recurrsive_convert_type(func)
  schema["$defs"][func["name"]] = {
    "type": "object",
    "description": func.get("description", ""),
    "properties": {
      func["name"]: new_func["parameters"]
    },
    "required": [func["name"]],
    "additionalProperties": False
  }
  schema["$defs"]["tools"]["oneOf"].append({ "$ref": "#/$defs/{}".format(func['name'].replace('~', '~0').replace('/', '~1')) })
  schema["oneOf"] = [
    {
      "type": "array",
      "description": "Calling multiple tools in a array.",
      "items": {
        "$ref": "#/$defs/tools"
      },
      "minItems": 2
    },
    {
      "$ref": "#/$defs/tools"
    },
    {
      "type": "string",
      "description": "If none of the function can be used, point it out here. If the given question lacks the parameters required by the function, also point it out here."
    }
  ]
  jsonschema.Validator.check_schema(schema)
return schema
\end{minted}

\onecolumn

\end{document}